\documentclass[journal]{IEEEtran}
\usepackage[T1]{fontenc}
\usepackage[utf8]{inputenc}

\usepackage{amsmath,amssymb,amsthm,mathtools}
\usepackage{graphicx,booktabs,array,tabularx}
\usepackage{xcolor}
\usepackage{cite}
\usepackage[hyphens]{url}
\usepackage[hidelinks]{hyperref}
\usepackage{orcidlink}

\theoremstyle{plain}

\theoremstyle{definition}

\newcommand{\TODO}[1]{\textcolor{red}{[TODO: #1]}}
\newcommand{\NModTotal}{47,885}
\newcommand{\NModCovered}{47,780}
\newcommand{\NDeclReal}{871,784}
\newcommand{\NZeroCross}{454,574}
\newcommand{\NSameFileOnly}{438,545}
\newcommand{\NTrueZero}{16,029}
\newcommand{\PFalseZero}{96.5\%}
\newcommand{\PTrueZero}{1.8\%}

\title{``Not Used Elsewhere'' Is Not ``Unused'' in a Lean Formalization of the Riemann Hypothesis:\\
Reverse Lookup Against the Compiled Index, Based on Eduinformatics}

\author{\mbox{Kunihiko~Takamatsu~\orcidlink{0000-0003-4172-7612}}, \mbox{Katsuhiko~Murakami~\orcidlink{0000-0002-1923-4259}}, \mbox{Sayaka~Matsumoto~\orcidlink{0000-0002-4556-0744}}, \mbox{Tomohiro~Akiyama~\orcidlink{0000-0002-6720-8995}}, \mbox{Shimpei~Matsumoto~\orcidlink{0009-0005-6587-7852}}, \mbox{Katsuyuki~Kunida~\orcidlink{0000-0002-6346-2638}}, \mbox{Satoshi~Takahashi~\orcidlink{0000-0001-5573-7841}}, \mbox{Tokuro~Matsuo~\orcidlink{0009-0000-3551-8799}}, \mbox{Sachio~Hirokawa~\orcidlink{0000-0002-8050-1109}}, \mbox{Kazuya~Tsuruta~\orcidlink{0009-0004-7386-2290}}, \mbox{Ryosuke~Kozaki~\orcidlink{0000-0001-8428-9371}}, \mbox{Shuto~Nishida}, \mbox{Koma~Takamatsu~\orcidlink{0009-0001-8144-6797}}, \mbox{Ford~Lumban~Gaol~\orcidlink{0000-0002-5116-5708}}, and \mbox{Masao~Mori~\orcidlink{0000-0002-2883-8943}}
\thanks{Kunihiko Takamatsu, Sayaka Matsumoto, and Masao Mori are with the Institute of Science Tokyo, Tokyo, Japan. K. Murakami is with The University of Tokyo, Tokyo, Japan. T. Akiyama is Specially Appointed Associate Professor, Uehiro Research Center for Japan Environmental Studies, Research Institute for Humanity and Nature, Japan; Associate Professor (co-appointed), Global Environmental Studies Program, Graduate University for Advanced Studies, SOKENDAI, Japan; Project Researcher, Department of Multi-Disciplinary Sciences, Graduate School of Arts and Sciences, The University of Tokyo, Japan; Visiting Professor, Graduate School of Information Technology, Kobe Institute of Computing, Japan; Visiting Professor, Miyazaki International University, Japan; and Visiting Researcher, Institute for Contemporary Chinese Studies, Aichi University, Japan. Shimpei Matsumoto is with Hiroshima Institute of Technology, Hiroshima, Japan. K. Kunida and T. Matsuo are with Fujita Health University, Aichi, Japan. S. Takahashi is with The University of Electro-Communications, Tokyo, Japan. S. Hirokawa is with Kyushu University, Fukuoka, Japan, and Monolithic Design Co., Ltd., Fukuoka, Japan. K. Tsuruta is with Monolithic Design Co., Ltd., Fukuoka, Japan. R. Kozaki and S. Nishida are with International Pacific University, Okayama, Japan. Koma Takamatsu is with Kwansei Gakuin University, Hyogo, Japan. F. Lumban Gaol is with Binus University, Jakarta, Indonesia. Corresponding author: Kunihiko Takamatsu (ORCID 0000-0003-4172-7612).}}

\begin{document}
\maketitle
\sloppy

\begin{abstract}
On 20 November 2026 a Lean~4 proof of the Riemann Hypothesis of about 175,000 files was released, and some 200 volunteers began to consolidate and prune it. Deciding that a declaration is unused, and can be deleted, is a reverse lookup: it is only as reliable as the forward links it is built from. From the compiled files of the consolidated branch (commit e497a66) we built a forward and reverse index of all declarations, covering \NModCovered{} of \NModTotal{} modules. Of the \NZeroCross{} declarations that no other file uses, \PFalseZero{} are used inside their own file, so ``not used elsewhere'' badly overstates ``unused''; a deletion based on it broke the build and had to be partly reverted. Restoring the pre-consolidation partition of 174,793 files while holding the use relation fixed leaves this share unchanged (96.7\% on the mapped declarations), so the false zeros are not an artifact of the consolidation: helper declarations already sat in the files of their users. We then asked how much a capture--recapture estimate built from source text alone can say about use. When the compiled index serves both as one capture and as the ground truth, the ratio of the Lincoln--Petersen estimate to the truth equals the reciprocal of the text matcher's precision by construction; we make this identity explicit, and report what the comparison still shows: how low the precision of plausible text matchers is, how recall differs between same-file and cross-file uses, and which uses text cannot see at all. We give a procedure for safe deletion based on the compiled index.
\end{abstract}

\section{Introduction}\label{sec:intro}

On the evening of 25 November 2026 (JST), a pull request by one of the authors removed eight declarations from a volunteer-maintained Lean~4 repository, on the grounds that they were unused (pull request \#1416). Three minutes after it was merged, a second pull request restored six of them (\#1417)~\cite{repo_ra}. The check that preceded the deletion had searched for uses not only in other files but also in the file that declares each name. It nevertheless missed a use of \texttt{thZero} because the name occurred after a dot, as \texttt{V.thZero}, and the textual search did not match it there. A declaration that was in use was therefore deleted, and the build broke. The episode is small, but it shows the question this paper addresses: when a tool reports that a declaration is ``unused'', how far can that report be trusted, and what does the answer depend on?

\paragraph{The artifact}
On 20 November 2026, Levent Alpöge released, together with Claude, a Lean~4 development of 175,187 files and 17,374,215 lines whose final theorems are Mathlib's statements of the Riemann Hypothesis and of the corresponding statements for Dirichlet $L$-functions~\cite{alpoge_archive}. On the next day the Lean FRO and the Mathlib Initiative reported that the statements are Mathlib's own and that the proof is accepted by the Lean kernel~\cite{leanfro_blog}. The artifact was written largely by an AI system, and no human has read all of it. Kernel acceptance settles whether the final theorems follow from the axioms; it does not tell a reader which parts of the development carry the argument, which parts are redundant, or how the code could be reduced to something a person can study. For an artifact of this size written by AI, these questions must themselves be answered with tools, and the reliability of those tools becomes part of the verification.

\paragraph{The volunteer consolidation}
Within days, a community of about two hundred GitHub contributors began to rebuild, reduce and port the archive~\cite{repo_lc,repo_ra,repo_replay,takamatsu2026review}. Pull request \#120 in \texttt{leanprover-community/riemann-hypothesis}, opened on 22 November from the branch \texttt{opt/collapse-and-consolidate-4.36} of \texttt{rohanarun/riemann-hypothesis}, collapsed 28,169 thin wrapper files and bundled 93,601 small modules into 20,258, reducing the number of \texttt{.lean} files by 67.9\%~\cite{pr120}. Many contributors then fixed the build errors caused by this restructuring, and many pull requests removed declarations that appeared to be unused. Between 06:30 JST on 25 November and the last merge at 02:18 JST on 26 November, 178 pull requests by 18 contributors were merged on this branch, with a median of six minutes from opening to merge. The working rule for deletion was, in effect, ``no other file uses it, so it can be removed''. Repeated fixes to the same file also often touched the lines of the previous fix (81 of 204 consecutive fix pairs, 39.7\%, in the commits up to 25 November 06:13 UTC), part of which was a back-and-forth between two choices.

All numbers in this paper refer to commit \texttt{e497a66} (25 November 2026, 17:55 UTC), the last commit on that branch. Merging stopped there, and community work moved to \texttt{leanprover-community/riemann-hypothesis}~\cite{repo_lc}. The snapshot is therefore fixed, and every count reported below can be reproduced from it.

\paragraph{Deletion as a reverse lookup}
Deciding that a declaration is unused is a reverse lookup: from the forward relation ``declaration $a$ uses declaration $b$'', collected over the whole development, one asks for each $b$ whether any $a$ points to it. A reverse lookup is a transpose and is computed mechanically, but its answer is only as good as the forward relation it is built from. In earlier work we showed, on citation graphs of mathematics textbooks, that missing forward links turn into false ``never cited'' statements, gave a closed-form correction under random omission, and showed that the uniform correction fails when omission is biased by link type~\cite{takamatsu2026coverage}. The deletions in the consolidation are an instance of the same problem at a much larger scale. A forward relation restricted to uses from other files is a biased sample of all uses: it omits, by construction, every use inside the declaring file, and a textual search that also scans the declaring file omits whatever its pattern does not match, as in \texttt{V.thZero}.

\paragraph{Origin of the approach}
The approach comes from Eduinformatics, an interdisciplinary field proposed in 2018 that integrates education and informatics~\cite{edu2018}. A recurring theme of Eduinformatics is to collect primary records exhaustively and to analyse them directly, rather than through aggregated indicators that can hide what the records contain. Forward and reverse lookup over curricula and textbooks are a basic instance of this practice~\cite{takamatsu2026coverage}. Here we apply the same practice to a machine-written proof: we collect the forward relation exhaustively from the compiled files, transpose it, and use the result both to decide deletions and as ground truth against which cheaper estimates can be checked.

\paragraph{Contributions}
Several ingredients of this study are not new. Capture-recapture has been used to estimate the number of defects remaining after software inspections since 1992~\cite{eick1992}; the inflation of capture-recapture estimates by misidentification is well known in ecology~\cite{link2010}; and three groups have already counted the declarations of this proof that are reachable from its main theorems~\cite{zhuk9581,yth4377,schiller3750}. We build on these and claim three new results.
\begin{enumerate}
\item We measure how far the volunteers' working rule is from actual use. From the compiled files of the consolidated branch we built a forward and reverse index of \NDeclReal{} declarations in \NModCovered{} of \NModTotal{} modules. Of the \NZeroCross{} declarations that no other file uses, \NSameFileOnly{} (\PFalseZero) are used in their own file, and only \NTrueZero{} (\PTrueZero{} of all declarations) have no recorded use (Fig.~\ref{fig:falsezero}). This share of false zeros, declarations that no other file uses but their own file does, is not an artifact of the consolidation: holding the compiled use relation fixed and restoring the pre-consolidation file partition (174,793 \texttt{.lean} files), the share on the 452,153 declarations without cross-file use that could be mapped (99.47\%) is 96.73\% under the current partition and 96.72\% under the old one, and only 211 of the 437,349 false zeros (0.048\%; at most 0.51\% under every mapping we tried) had a cross-file use under the old partition (Section~\ref{sec:3-filegrouping}). Nor does it depend on how compiler-generated auxiliary constants are attributed to declarations: under a narrower and a broader rule the share is 95.8\% and 96.7\%. We link this gap to deletions that had to be partly reverted, one of which broke the build.
\item We compare text-based capture-recapture with a complete ground truth at scale, rather than with seeded defects. On 1,000 modules chosen at random from the compiled index, with 88,003 compiled uses as ground truth, a simple text matcher (precision 81.8\%, recall 16.1\%) yields an estimate 1.22 times the true number of cross-file uses, whereas a matcher that resolves \texttt{open} and \texttt{namespace} (precision 87.4\%, recall 45.2\%) yields 1.14 times (Fig.~\ref{fig:cr}). These ratios agree with the reciprocals of the precisions (1/0.818 = 1.22, 1/0.874 = 1.14), and this is not an empirical finding: when the compiled index serves both as one capture and as the ground truth, the ratio of the Lincoln--Petersen estimate to the truth equals the reciprocal of the matcher's precision by construction, so recall does not enter it. What the comparison adds is how far the measured (Chapman) ratios depart from this identity, and how far below 1 the precision of a plausible text matcher is in practice.
\item We give a deletion procedure based on the compiled index, which counts same-file uses, and report the pull requests made with earlier, weaker versions of it (cross-file uses from the compiled index plus a text search for same-file uses), including those that were accepted, reverted or left pending (Table~\ref{tab:prs}). On the snapshot branch, 18 pull requests of the first author were merged (\#1415--\#1433; \#1423 is by another contributor): 16 deletions, including \#1416, whose same-file check was a textual search, and 2 restorations (\#1417 restored six declarations removed by \#1416; \#1425 restored five declarations removed by \#1420). \#1420 did not break the build; \#1425 restored declarations that \#1420 had removed outside its intended scope (Section~\ref{sec:index:case}). In addition, 14 pull requests (\#1434--\#1449, excluding \#1440 and \#1441, which are not the first author's) remained open when merging on the branch stopped.
\end{enumerate}

\paragraph{Outline}
Section~\ref{sec:background} reviews false ``never cited'' counts under incomplete forward links and their correction, capture-recapture in software inspection and ecology, and the first week of research on the release. Section~\ref{sec:index} describes the compiled index, the share of declarations that are used only in their own file, a comparison of that share under the pre-consolidation file partition (Section~\ref{sec:3-filegrouping}), and a case study of deletions that were reverted. Section~\ref{sec:cr} evaluates text matchers against the compiled index, which serves both as one capture and as the ground truth. Section~\ref{sec:discussion} discusses why false matches inflate the estimate, what lies outside the scope of a capture, the instability of the estimate when the overlap is small, and a rule for safe deletion. Section~\ref{sec:limitations} states the limitations, the public input, and the use of generative AI in this work.

\section{Background and related work}\label{sec:background}

\subsection{False ``never cited'' counts under incomplete forward links}\label{sec:bg-falsezero}

Our approach rests on an earlier study of forward and reverse lookup~\cite{takamatsu2026coverage}.
Recording ``$A$ uses $B$'' indexed by $A$ gives a forward lookup; reading the same records indexed by $B$ gives a reverse lookup.
The reverse lookup is a transpose and is computed mechanically, but its results are meaningful only if the forward lookup is exhaustive.
In particular, the judgement that an item is ``never cited'' (in the present setting: that a declaration is unused) mixes true zeros with false zeros caused by missing links.

That study formalised this as follows.
If each link of the complete forward lookup is retained independently with probability $q$ (the coverage), the expected number of false zeros is $\sum_{v:\,k_v>0}(1-q)^{k_v}$, where $k_v$ is the true number of links to item $v$; items with a single true use are the most likely to be wrongly judged unused (Proposition~1 of~\cite{takamatsu2026coverage}).
If $q$ is known, the estimator $\hat z_0=\sum_v s^{k'_v}$ with $s=-(1-q)/q$, where $k'_v$ is the observed count, is unbiased for the true number of zeros for every $0<q<1$ and has a closed-form variance; the variance contribution of heavily used items shrinks only when $q>1/2$ (Proposition~2). The estimator has the form of a Good--Toulmin estimator~\cite{good1956}.
When links are of several types with different coverages $q_\tau\in(0,1]$, the correction is applied per type (Corollary~1).
On a citation graph of mathematics textbooks, the correction recovered the true count under uniform omission (at $q=0.9$, a naive count of 889.1 against a true value of 821, corrected to 821.4), but applying the uniform correction to a biased omission that drops cross-book citations gave 494.3, worse than the uncorrected count of 878~\cite{takamatsu2026coverage}.

The working rule examined in this paper, ``no use from other files, so the declaration can be deleted'', is a reverse lookup on a forward lookup that keeps cross-file uses and omits all same-file uses.
In the terms above, this is the extreme case of biased omission: the coverage of the same-file type is $q_{\text{same}}=0$, which lies outside the range of Corollary~1, so no reweighting of the observed counts can recover the missing uses.
They have to be measured.
Moreover, which uses count as ``cross-file'' is fixed by how declarations are grouped into files, not by the use relation itself; consolidating files turns cross-file uses into same-file uses without changing any use.
Section~\ref{sec:index} measures the false-zero share of this rule on the compiled index (Fig.~\ref{fig:falsezero}).
As suggested by T.~Akiyama, Section~\ref{sec:3-filegrouping} also recomputes the shares of Fig.~\ref{fig:falsezero} under the pre-consolidation file partition with the use relation held fixed, which separates how much of the share is due to file grouping alone.

\subsection{Capture--recapture in software inspection and in ecology}\label{sec:bg-cr}

Experiment~4 of~\cite{takamatsu2026coverage} treated two mechanical extractions of the same citation graph (citations by label and by name) as two captures and computed the two-list Chapman estimate $\hat N=(n_1+1)(n_2+1)/(m+1)-1$~\cite{chapman1951,chao2001}, where $n_1$ and $n_2$ are the numbers of links found by each extraction and $m$ the number found by both.
Only 21.4\% of the citations by label were also found by name, and the implied coverage of 27--35\% could be read only as a conditional guideline: the two extractions are negatively dependent, because an author needs to cite either by label or by name, and name matching also yields false links; both inflate $\hat N$, whereas heterogeneous catchability works in the opposite direction~\cite{chao2001}.
That study had no ground truth against which to check the estimate.
The present paper supplies one: the compiled index of the Lean artifact records every use that the elaborator resolved.

Capture--recapture has been used to estimate the number of defects remaining after a software inspection since Eick et al.~\cite{eick1992}, who treated the defects found by each inspector as a capture.
Later work compared many estimators and model assumptions for inspection data~\cite{briand2000}.
These evaluations used data of modest size, and the units being counted (defects) are identified by human judgement rather than by a matcher.
Eick et al.\ applied their method to 13 review sessions of feature requirements or feature designs, for which the true number of defects was not known; a longitudinal study tracing these features was in progress~\cite{eick1992}.
Briand et al.\ used two experiments in which four requirements documents were inspected, with data from 12 and 13 professionals analysed; the 15 to 29 defects of each document, made while the document was written or detected in later phases of development, had been reintroduced (seeded) before the experiments, so the true count was known~\cite{briand2000}.

In ecology, the effect of misidentification on capture--recapture is well known.
When individuals are identified from genetic samples, genotyping errors create ``ghost'' individuals that are never recaptured; they enter the count of marked animals but not the overlap, and the population is overestimated~\cite{creel2003,lukacs2005}.
In the two-list setting, if one list is the complete ground truth with $C$ items (no false and no missing items), and the other contains $T$ items of which $M$ are correct matches, the Lincoln--Petersen estimate is $CT/M$, so its ratio to $C$ is $T/M$, the reciprocal of the precision of the second list; this is an algebraic identity, not an empirical finding.
False matches by a source-text matcher play the role of ghost individuals.
Section~\ref{sec:cr} uses this identity as a reference line (Fig.~\ref{fig:cr}).

\subsection{Declarations reachable from the main theorems}\label{sec:bg-reach}

The Lean~4 development released on 20 November 2026 by Alpöge with Claude~\cite{alpoge_archive} was soon examined by several verification studies.
At the level of the whole artifact, three groups computed the set of declarations reachable from the five root theorems: 1,478,222~\cite{zhuk9581}, 1,478,243~\cite{yth4377} and 1,478,219~\cite{schiller3750}.
Schiller reports two independent transitive traversals of the compiled environment that agree~\cite{schiller3750}; yth also reports the effect of early community pruning on the number of source declarations~\cite{yth4377}.
Dependency maps for human readers, at the level of tens to hundreds of declarations, were given by Gifford~\cite{gifford9970} and Zhuk~\cite{zhuk9581}.
All of these counts refer to the original archive, not to the consolidated branch studied here, and they answer a different question.
Reachability from the main theorems decides which declarations the proof needs; it does not tell a volunteer whether deleting a particular declaration will break the build, because a declaration that the proof does not need can still be used by another declaration in the same or another file.
For that question the relevant relation is the full use relation among source declarations, including same-file uses, which is what our index records.

\subsection{The first week of research}\label{sec:bg-review}

A review of the first week of preprints on the release~\cite{takamatsu2026review} found at least 89 documents within six days, none peer reviewed, of which nine were verification studies.
Those studies agree that the top-level theorems are Mathlib's statements, that only the three standard axioms are used, and that the archive builds.
The review also found that no paper examines the volunteer process that consolidated and pruned the artifact: how about two hundred contributors' pull requests were reviewed and merged, how often fixes were undone, and how a declaration came to be judged unused.
One such restructuring, pull request \#120, reduced the number of \texttt{.lean} files from 149,396 to 47,884~\cite{pr120}.
The present paper addresses the last of these questions.

\section{The compiled index}\label{sec:index}

This section describes how we built a forward and reverse index of the declarations of the consolidated branch from its compiled files, what the index says about the volunteers' working rule ``no use from other files, so delete'', and what happened when deletions were made on the basis of a weaker check. All index numbers refer to commit \texttt{e497a66} of branch \texttt{opt/collapse-and-consolidate-4.36} of \texttt{rohanarun/riemann-hypothesis}.

\subsection{Construction from compiled files}\label{sec:index:build}

Lean~4~\cite{demoura2021lean4} stores the result of elaborating each module in a compiled file (\texttt{.olean}, with \texttt{.olean.private} for private parts). Every declaration is stored there as a kernel term, so the constants that a declaration's type and value refer to can be read without parsing the source text, and without the ambiguities of \texttt{open}, \texttt{namespace}, dot notation and notation that a text search has to resolve. We wrote a short Lean program that imports a list of modules, walks the constants stored for each module of the project (Mathlib and other dependencies are skipped), and for each constant collects the set of constants it uses. For each module it writes three kinds of rows:
\begin{itemize}
\item \textbf{D} (declaration): module $m$ declares name $n$;
\item \textbf{U} (use from other files): module $m$ uses name $n$, and $n$ is not declared in $m$;
\item \textbf{S} (same-file use): declaration $a$ uses declaration $b$, both declared in the same module, $a\neq b$.
\end{itemize}
The U rows are the forward links between files; transposing them gives the reverse lookup ``which files use $n$''. The S rows keep the using declaration as well as the used one, so that a declaration's uses by its own auxiliary constants can be removed later. Extensions (attributes, simp sets and other environment extensions) are not loaded, so the index records only references that occur in kernel terms.

\paragraph{Auxiliary names}
The elaborator generates many constants that no person wrote: proof, simp and match auxiliaries (\texttt{\_proof\_$k$}, \texttt{\_simp\_$k$}, \texttt{match\_$k$}, \texttt{match\_$k$\_$l$}), equation lemmas (\texttt{eq\_$k$}, \texttt{eq\_def}, \texttt{eq\_unfold}), the recursors and related constants of inductive types and their constructors (\texttt{rec}, \texttt{recOn}, \texttt{casesOn}, \texttt{below}, \texttt{brecOn}, \texttt{noConfusion}, \texttt{injEq}, \texttt{inj}, \texttt{ctorIdx}, \texttt{ctorElim}, \texttt{splitter}, \texttt{sizeOf\_spec}), notation helpers (\texttt{\_aux\_\ldots}), and other names beginning with an underscore. The auxiliaries of a private declaration carry a prefix \texttt{\_private.\textit{M}.0.}, where \textit{M} is the module (for example \texttt{\_private.\textit{M}.0.foo.\_proof\_1}). We map every name to its \emph{root}, the prefix before the first such component, keeping or dropping the prefix \texttt{\_private.\textit{M}.0.} so that the root is a declared name; a use of an auxiliary constant is counted as a use of its root, so every auxiliary constant is folded into the user-written declaration it belongs to. A declaration is counted as \emph{real} if it is its own root. An S row $(a,b)$ is counted as a same-file use of $b$ only if the root of $a$ differs from $b$ and from the root of $b$; otherwise a theorem would be counted as ``used'' merely because its own equation lemma or proof auxiliary refers to it. An earlier version of this rule missed the auxiliaries of private declarations and several of the constructor helpers listed above. The main claim does not depend on these details: in an independent audit, the share of false zeros reported in Section~\ref{sec:index:falsezero} is 95.8\% under that narrower rule and 96.7\% under a broader rule that also treats structure constructors (\texttt{mk}), the \texttt{ext\_iff} lemmas generated by \texttt{@[ext]} and the names generated for notation as auxiliary.

\paragraph{Coverage}
The index covers the modules that compile locally. These are \NModCovered{} of the \NModTotal{} modules of the branch. The remaining 105 modules contain no declarations of their own: 104 consist only of \texttt{import} lines, and the last is \texttt{Challenge}, the statement file that is built against Mathlib alone and contains three definitions. Missing modules can only add uses, never remove them, so with respect to the uncovered part the counts of unused declarations below are upper bounds, in the sense of~\cite{takamatsu2026coverage} (a forward lookup $F'\subseteq F$ gives $k'_v\le k_v$, the observation preceding its Proposition~1): an incomplete forward lookup produces false zeros, never false non-zeros. The local build used the pinned toolchain \texttt{leanprover/lean4:v4.36.0} on one Apple M3 Ultra machine (32 cores, 512\,GB memory).

\subsection{How far ``not used elsewhere'' is from ``unused''}\label{sec:index:falsezero}

\begin{figure}[t]
\centering
\IfFileExists{figs/fig_falsezero.pdf}{\includegraphics[width=\linewidth]{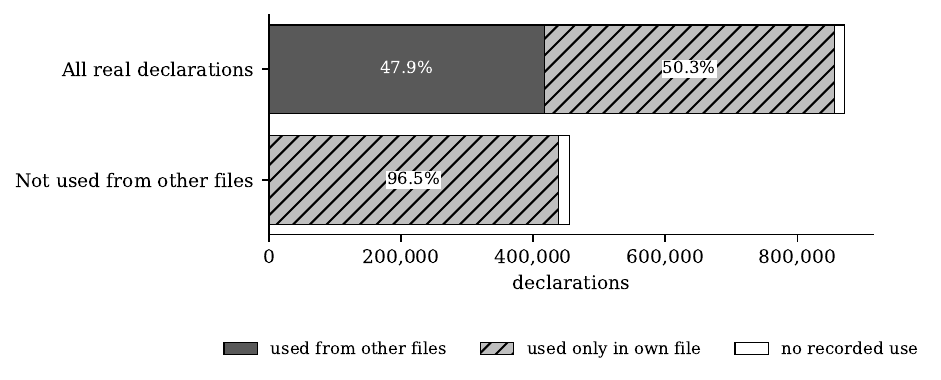}}{\fbox{\TODO{fig\_falsezero.pdf}}}
\caption{Real declarations in the compiled index by recorded use. Top: all \NDeclReal{} real declarations. Bottom: the \NZeroCross{} declarations with no use from other files, of which \PFalseZero{} are used in their own file (hatched) and only the rest have no recorded use (white). Covered modules: \NModCovered{} of \NModTotal{}.}
\label{fig:falsezero}
\end{figure}

The index contains \NDeclReal{} real declarations. Of these, \NZeroCross{} have no U row, that is, no use from any other covered file; this is the set that the working rule ``no use from other files, so delete'' would mark as deletable. But \NSameFileOnly{} of them (\PFalseZero) are used inside their own file. Only \NTrueZero{} declarations, \PTrueZero{} of all real declarations, have no recorded use at all. Figure~\ref{fig:falsezero} shows the breakdown.

In the terms of~\cite{takamatsu2026coverage}, ``no use from other files'' is a reverse lookup built from a forward lookup that omits one whole type of edge, the same-file uses. It is not a random omission of the kind treated in Propositions~1 and~2 of that paper, where each link is kept independently with the same probability; the omission is complete for one type and absent for the other. The share of false zeros is therefore not a sampling effect that a larger sample would reduce, and the estimator of Proposition~2 cannot correct it. Corollary~1 of~\cite{takamatsu2026coverage} treats omission that differs by type of link, but it assumes that each type is retained with a positive probability; here same-file uses are retained with probability zero. The only remedy is to add the missing type of link, which is what the S rows do.

Most of these false zeros are auxiliary declarations, helper lemmas and definitions, some of them private, that are used by a parent declaration in the same file. Section~\ref{sec:3-filegrouping} asks whether the large share is instead an artifact of how the declarations are grouped into files.

\subsection{Does file grouping explain the false zeros?}\label{sec:3-filegrouping}

Consolidation bundled many small source files into larger ones (pull request \#120 of \texttt{leanprover-community/riemann-hypothesis}), so a helper lemma that used to be imported from another file may now sit next to its user. Whether a use counts as ``from another file'' depends on how the declarations are grouped into files, not only on the use relation itself, and the large share of false zeros could therefore be an artifact of consolidation. T.~Akiyama proposed a direct test, which we carried out: recompute the share of false zeros under the pre-consolidation file partition while holding the use relation fixed.

\paragraph{Method}
The use relation is that of the compiled index at \texttt{e497a66}; only the partition into files is changed, to that of commit \texttt{40e56bab02} before consolidation, which has 174,793 \texttt{.lean} files. We mapped each declaration to its file at \texttt{40e56bab02} in two ways. First, we parsed the old source text without compiling it, following \texttt{namespace}, \texttt{section} and \texttt{mutual} blocks and modifiers such as \texttt{private}, and attached auxiliary names to their parent by the longest matching prefix. Second, the consolidated files at \texttt{e497a66} are divided by markers \texttt{-- bundled from} followed by the old module name (93,598 markers in 20,259 bundled files), and we assigned each declaration to the segment in which it lies. Of the declarations mapped by both methods, 99.985\% are mapped to the same old file. Consolidation was essentially a pure merge: only 60 of 120,637 old files (0.05\%) have declarations in more than one current module. A declaration with no use from other files counts as having a use from another file under the old partition if some declaration that uses it in its own file lies in a different old file. The main estimate uses the parsed mapping, with mappings that contradict the markers left unmapped; it covers 452,153 of the \NZeroCross{} declarations with no use from other files (99.47\%).

\paragraph{Result}
On this subset the share of false zeros is 96.73\% under the current partition (437,349 of 452,153) and 96.72\% under the pre-consolidation partition (437,138 of 451,942). Only 211 of the 437,349 false zeros (0.048\%) would have a use from another file under the old partition. Counting every declaration whose user could not be mapped as such a case gives at most 931 (0.21\%), the largest such upper bound over all mappings is 0.51\%, and the marker mapping alone gives 217. Because the use relation is fixed at \texttt{e497a66}, it omits uses from the 53,966 old files that consolidation removed, many of them thin wrappers around other theorems. If every declaration with no use from other files whose name (its last two components) occurs in one of these removed files is counted as used from another file, at most 2,843 false zeros (0.65\%) change, and the share is 96.83\%; it rises slightly, because true zeros whose names occur in the removed files are counted as used from other files as well. Of the \NSameFileOnly{} false zeros, 60,349 (13.8\%) are private declarations.

The false zeros are therefore not an artifact of consolidation. They reflect that helper declarations, lemmas and definitions used by a parent declaration, live in the same file as their users, as they already did before consolidation. The test has limits: declarations deleted, renamed or changed between \texttt{40e56bab02} and \texttt{e497a66} are reflected on the old side only through the name-based sensitivity check above, the parser is approximate, and nothing was compiled for this comparison.

\paragraph{Implication}
The gap between ``not used elsewhere'' and ``unused'' is thus a property of how this proof is written, not of how its files happen to be grouped. A theorem and the helpers that serve it are written side by side, so under either partition, before or after consolidation, a check limited to other files misses almost all of the uses that keep a helper alive. Merging or splitting files moves only a tiny fraction of uses (here at most 0.51\%) across the line between same-file and cross-file. For deletion this means that no reorganisation of files makes a cross-file check safe: a safe check must include same-file uses whatever the file layout (Section~\ref{sec:disc-rule}). We expect the same in other large Lean developments written in this style, but we have measured only this snapshot.

\subsection{Case study: deletions that were reverted}\label{sec:index:case}

Volunteers' pull requests rarely state how ``unused'' was established; the merged removals with ``unused'' in the title that we found in \texttt{leanprover-community/riemann-hypothesis} say only that the declaration ``has no callers'' (\#6792, \#7387, \#7532) or that it is not reachable from \texttt{Main.lean}, \texttt{Statements.lean} or \texttt{Challenge.lean} (\#6269). The first author also made deletions, and two of them were reverted within an hour (3 and 50 minutes after the merge). Both are recorded in the merged history at the snapshot.

\paragraph{\#1416 $\to$ \#1417 (missed same-file uses)}
Pull request \#1416 removed eight declarations. The check behind it took cross-file uses from the compiled constants (D and U rows only; S rows were not yet recorded) and covered uses inside the declaring file with a text search. That text search skipped occurrences preceded by a dot, so it missed fully qualified names and dot notation in the same file. Six of the eight declarations were in fact used later in their own files, and their removal broke five modules. One example: in \texttt{AutomorphicMembership/Bundle0005.lean} the declaration \texttt{\ldots VacIn.thZero} is used as \texttt{V.thZero}, and \texttt{thZero} in turn uses \texttt{thMM}, which was removed together with it; the search missed the use \texttt{V.thZero}. In addition, \#1416 was submitted before the local build of the affected modules had finished, and it was merged before the repository's compile check ran. Pull request \#1417, merged three minutes later, restored the six declarations as an exact reverse of the corresponding part of \#1416; the other two deletions built and were kept. This is precisely the false-zero case of Section~\ref{sec:index:falsezero}: the six declarations had no use from other files but were used in their own file.

\paragraph{\#1420 $\to$ \#1425 (an unchecked edit carried into a branch)}
Pull request \#1420 was meant to change one file. The script that prepares one pull request per file re-checks each candidate on the current head just before opening it. For another file, \texttt{IsUnitaryOfOrder.lean}, that re-check failed because a qualified suffix of one of the candidates (\texttt{HChar.resInf\_apply}) occurs as text in two other files, so no pull request was opened for it; but the edit was left in the working tree and carried into the branch of \#1420. Pull request \#1425 restored the five declarations. This failure was in the tooling, not in the index. In the compiled index at the snapshot we find no use of any of the five declarations from other files, although both files with the textual match are covered; within their own file the only recorded uses are of the auxiliary proof constant \texttt{etaAlg.\_proof\_1}, by \texttt{Mu0Exists} (itself one of the five) and by \texttt{EtaTExists} and \texttt{DeltaExists}, which our counting rule treats as same-file uses of \texttt{etaAlg}; the textual occurrence was a false match of the kind measured in Section~\ref{sec:cr}. These same-file uses are not genuine dependencies on \texttt{etaAlg}: Lean shares an identical proof term among declarations by abstracting it as an auxiliary constant under the first declaration that needs it, and when \texttt{etaAlg} is removed the term is abstracted under another declaration. On 27~November 2026 we compiled \texttt{IsUnitaryOfOrder.lean} with Lean as of the merge commit \texttt{94167b2} (after \#1420) and as before \#1420; both versions compile without errors. Both were compiled against the dependencies of the snapshot \texttt{e497a66}, not those of \texttt{94167b2}. Within this limit, we find no evidence that \#1420 broke the build: \#1425 restored declarations removed outside the intended scope of \#1420, rather than repairing a broken build.

The two cases fail in opposite directions. In the first, a text search missed real uses (a recall failure) and a used declaration was deleted. In the second, a text search found a use that does not exist (a precision failure); it stopped the deletion correctly in intent, and the unintended deletion came from a process step around it. After \#1417, every deletion was checked against the compiled index for uses from other modules (S rows were not yet recorded), by a text search for the short name in the declaring file that counts qualified and dot-notation occurrences, and by a search for the qualified suffix in all other files and in the uncompiled modules; declarations carrying attributes such as \texttt{@[simp]}, \texttt{@[instance]} or \texttt{@[ext]} were excluded, because uses through attributes do not appear as references in kernel terms. Each edited module was built locally before submission.

\subsection{Pull requests of the first author}\label{sec:index:prs}

Table~\ref{tab:prs} lists the pull requests of the first author on the branch. Eighteen were accepted (merged) between 19:52 on 25~November and 02:18 on 26~November (JST): \#1415--\#1433, except \#1423, which is by another contributor. Sixteen of them were deletions, including \#1416, whose check for same-file uses was a text search, and two were the restorations above: \#1417 restored six declarations removed by \#1416, and \#1425 restored five declarations removed by \#1420. The first three deletions (\#1415, \#1416, \#1418) removed unused declarations whose short names collide with live declarations in other namespaces, so that after bundling they only add candidates to name resolution; some of these names had already required disambiguation fixes. The later thirteen (\#1419--\#1433, excluding \#1423 by another contributor and the restoration \#1425) removed unused copies of declarations whose statement text is identical, apart from the name, to that of a used declaration elsewhere; these pairs were found by comparing the statements of 805,239 declarations, which gave 32,703 identical pairs. At the time of these pull requests the compiled data covered 39,476 modules. By their titles these pull requests removed 38 declarations, of which 6 were restored by \#1417, so 32 stayed removed; the five declarations restored by \#1425 were not among the intended deletions. When merging on the branch stopped, 14 further pull requests of the first author (\#1434--\#1449, excluding \#1440 and \#1441, which are not the first author's) remained open; community work moved to \texttt{leanprover-community/riemann-hypothesis}.

\begin{table*}[t]
\centering
\caption{Pull requests of the first author on branch \texttt{opt/collapse-and-consolidate-4.36} (times are JST, 25--26 November 2026).}
\label{tab:prs}
\footnotesize
\setlength{\tabcolsep}{4pt}
\begin{tabularx}{\textwidth}{@{}l>{\raggedright\arraybackslash}p{2.1cm}>{\raggedright\arraybackslash}X>{\raggedright\arraybackslash}p{2.6cm}@{}}
\toprule
PR & Merged & Change & Status \\
\midrule
\#1415 & 25 Nov 19:52 & Remove 5 unused declarations that collide with live namesakes & accepted \\
\#1416 & 25 Nov 21:01 & Remove 8 unused declarations that collide with live namesakes & accepted; 6 restored \\
\#1417 & 25 Nov 21:04 & Restore 6 declarations removed in \#1416 (used in their own files) & accepted (restoration) \\
\#1418 & 25 Nov 21:25 & Remove 10 unused declarations that collide with live namesakes & accepted \\
\#1419--\#1433$^{a}$ & 25 Nov 22:56 -- 26 Nov 02:18 & Remove 15 unused duplicate declarations, one file per PR & 13 accepted \\
\#1420 & 25 Nov 23:20 & (included above) carried an unchecked edit to a second file & accepted; edit restored \\
\#1425 & 26 Nov 00:10 & Restore 5 declarations removed by mistake in \#1420 & accepted (restoration) \\
\#1434--\#1449$^{b}$ & --- & Further deletions & 14 open, not merged \\
\bottomrule
\end{tabularx}

\smallskip
\raggedright\footnotesize $^{a}$ \#1419, \#1420, \#1421, \#1422, \#1424, \#1426, \#1427, \#1428, \#1429, \#1430, \#1431, \#1432, \#1433; 13 PRs removing 1 or 2 declarations each (\#1423 is by another contributor; \#1425 is listed separately). $^{b}$ \#1440 and \#1441 are not the first author's.
\end{table*}

\section{Evaluating text matchers against the compiled index}
\label{sec:cr}

\paragraph{Role of the compiled index}
In the experiment reported here the compiled index plays two roles at once: it is the first capture, and it is also the ground truth against which the estimate is judged.
The second capture is a text matcher run on the Lean source files.
Nothing in this section is therefore an estimate made without compiling; it is a measurement of how far an estimate that \emph{could} be made from text would be from the compiled answer, and of what in that gap can and cannot be seen from text.
As shown below, this design makes the headline ratios an algebraic identity in the precision of the text matcher.
We state this first so that the numbers are not read as an empirical discovery.

\paragraph{Unit of counting}
The unit is a distinct pair $(m,t)$, where $m$ is a using module (one source file) in the sample and $t$ is a declaration used by that module.
Occurrences are not counted: a declaration used many times in one file contributes one pair.
Auxiliary declarations generated by the elaborator (for example \texttt{match\_}$k$, \texttt{proof\_}$k$, \texttt{eq\_def}, \texttt{rec}, \texttt{casesOn}, including those under private declarations and constructor helpers) are mapped to the declaration they belong to, and only such root declarations are counted as $t$.
A pair is \emph{cross-file} if $t$ is declared in a module other than $m$, and \emph{same-file} otherwise.
In the compiled index a pair is recorded when $m$ uses $t$ directly, or when a declaration of $m$ uses $t$ and $t$ is not an auxiliary of that same declaration.

\paragraph{Sample and matchers}
We drew 1{,}000 modules at random (seed 20261126) from the \NModCovered{} modules of the compiled index built at commit \texttt{e497a66}; the source files were read at the same commit.
The compiled index records 88{,}003 pairs for these modules: 74{,}660 cross-file and 13{,}343 same-file.
We compared three text matchers of increasing care.
Matcher~A matches every pair of adjacent components of a dotted name in the text against the last two components of names declared in other files, and a short name that occurs at least twice against declarations of the same file.
Matcher~B first collects the prefixes introduced by \texttt{namespace} and \texttt{open} in the file and accepts a word only if some prefix plus the word is a declared name.
Matcher~C is B with the same-file rule of A added; for cross-file pairs it coincides with B, and for same-file pairs with A.
The scope handling of \texttt{open} and \texttt{namespace} in B is approximate (prefixes are pooled over the whole file).

\paragraph{The estimate is $1/\text{precision}$ times the truth}
Let $C$ be the number of pairs in the compiled index, $T$ the number of pairs proposed by the text matcher, and $M$ the number of pairs in both, all counted over the same scope.
The Lincoln--Petersen estimate \cite{lincoln1930,petersen1896} of the total is $\hat N_{\mathrm{LP}} = CT/M$.
Because $C$ is also taken as the truth,
\begin{equation}
  \frac{\hat N_{\mathrm{LP}}}{C} \;=\; \frac{T}{M} \;=\; \frac{1}{\text{precision of the matcher}},
  \label{eq:lp-identity}
\end{equation}
where precision is $M/T$, the share of proposed pairs that the compiled index confirms.
The recall of the matcher, $M/C$, cancels and does not enter the ratio.
The Chapman estimate \cite{chapman1951} $\hat N_{\mathrm{C}} = (C+1)(T+1)/(M+1)-1$ differs from $\hat N_{\mathrm{LP}}$ only by the added ones, which are negligible at these counts.
This relation was pointed out by T.~Akiyama while the paper was being drafted. We checked it, together with the exact difference $\hat N_{\mathrm{LP}}-\hat N_{\mathrm{C}}=(C-M)(T-M)/\bigl(M(M+1)\bigr)$ and its consequence that $\hat N_{\mathrm{C}}/C\le 1/\text{precision}$ whenever $M\le\min(C,T)$, by a formal proof in Lean~4 with Mathlib; the proofs use only the three standard axioms (\texttt{propext}, \texttt{Classical.choice}, \texttt{Quot.sound}), and the same file confirms by exact rational arithmetic that the two cross-file rows of Table~\ref{tab:cr} differ from $1/\text{precision}$ by less than $10^{-3}$.

\begin{figure}[t]
\centering
\includegraphics[width=\linewidth]{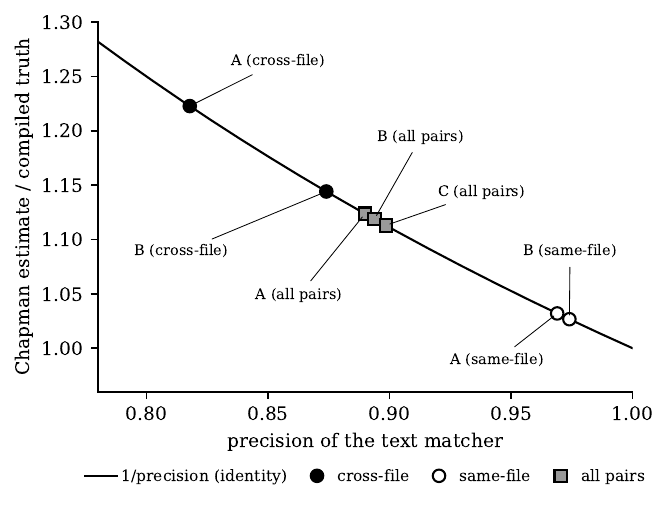}
\caption{Ratio of the capture--recapture estimate to the compiled truth against the precision of the text matcher, for 1{,}000 random modules. Solid line: $1/\text{precision}$, the identity \eqref{eq:lp-identity} that holds when the compiled index is both a capture and the truth. Points: Chapman estimates (filled circles: cross-file; open circles: same-file; squares: all pairs; labels A, B, C give the matcher).}
\label{fig:cr}
\end{figure}

Table~\ref{tab:cr} gives the counts.
For cross-file pairs, matcher~A has precision $12{,}025/14{,}704 = 81.8\%$, so by \eqref{eq:lp-identity} its estimate is $1.22$ times the truth; matcher~B has precision $33{,}711/38{,}572 = 87.4\%$ and gives $1.14$ times.
For same-file pairs the precision of every matcher is about 97\% (0.969 and 0.974), and the ratios are 1.03.
The Chapman ratios agree with $1/\text{precision}$ to the third decimal place in every cell.
Fig.~\ref{fig:cr} plots the ratio of the estimate to the truth against precision: the curve $1/x$ is the identity \eqref{eq:lp-identity}, and the Chapman values are drawn as points. At the scale of Fig.~\ref{fig:cr}, the Chapman estimates are indistinguishable from the curve; computed from the integers $C$, $T$, $M$ of Table~\ref{tab:cr}, all seven lie just below it, by at most $1.6\times10^{-5}$ (cross-file, matcher~A).
Our earlier statement that the error of the estimate ``is thus driven by precision (false matches), not by recall'' \cite{takamatsu2026review} is true, but by construction: in a design where one capture is the truth, it could not have come out otherwise.

\begin{table*}[t]
\centering
\caption{Capture--recapture on 1{,}000 random modules, with the compiled index as both first capture and truth. Unit: distinct (using module, used root declaration) pairs. $C$: pairs in the compiled index; $T$: pairs proposed by the text matcher; $M$: pairs in both. Ratio $=\hat N_{\mathrm{C}}/C$ (Chapman).}
\label{tab:cr}
\small
\begin{tabular}{llrrrrrrr}
\hline
Pairs & Matcher & $C$ & $T$ & $M$ & Recall & Precision & $1/\text{prec.}$ & Ratio \\
\hline
Cross-file & A & 74{,}660 & 14{,}704 & 12{,}025 & 0.161 & 0.818 & 1.22 & 1.22 \\
           & B (= C) & 74{,}660 & 38{,}572 & 33{,}711 & 0.452 & 0.874 & 1.14 & 1.14 \\
Same-file  & A (= C) & 13{,}343 & 13{,}404 & 12{,}988 & 0.973 & 0.969 & 1.03 & 1.03 \\
           & B & 13{,}343 & 9{,}531 & 9{,}283 & 0.696 & 0.974 & 1.03 & 1.03 \\
All        & A & 88{,}003 & 28{,}108 & 25{,}013 & 0.284 & 0.890 & 1.12 & 1.12 \\
           & B & 88{,}003 & 48{,}103 & 42{,}994 & 0.489 & 0.894 & 1.12 & 1.12 \\
           & C & 88{,}003 & 51{,}976 & 46{,}699 & 0.531 & 0.898 & 1.11 & 1.11 \\
\hline
\end{tabular}
\end{table*}

\paragraph{What the experiment still shows}
The identity fixes the ratio once precision is known, but it does not tell us what the precision is, nor how the two captures differ.
The data answer four questions that the identity leaves open.

First, how large the bias of a text-only estimate would be in this code base.
A matcher of the kind a volunteer would write in a few lines (matcher~A) proposes 14{,}704 cross-file pairs of which 18\% are false, because the last two components of a name are shared by many declarations in different namespaces.
Resolving \texttt{open} and \texttt{namespace} raises the precision to 87.4\% and the recall from 0.161 to 0.452.
This is the misidentification effect long known in ecology, where a mark that is read wrongly creates apparent new individuals and inflates the estimate \cite{link2010}; here a name that is resolved wrongly creates apparent new uses.

Second, the two captures do not see the same kinds of use with the same probability.
With matcher~A the recall is 0.973 for same-file pairs but 0.161 for cross-file pairs.
Two-list capture--recapture assumes that every unit has the same chance of being caught by a given list; pooling the two types violates this assumption, and the pooled ratio (1.12 for A) describes neither type.
We therefore report the estimate separately for each type of use, and recommend the same stratification whenever a text matcher is used on Lean sources.

Third, the Chapman correction does not matter here.
The overlaps are in the thousands (the smallest is $M=9{,}283$), so the small-overlap instability of the Chapman estimate, and of its normal-approximation interval, does not arise in this sample.
We do not report intervals: they quantify sampling variation under the capture--recapture model, whereas the error measured here is a systematic violation of that model.

Fourth, part of the compiled truth lies outside what text can see.
Uses created by elaboration---instance resolution, notation and macros, coercions, and declarations generated by \texttt{deriving} or by structure and inductive definitions---have no name in the source text that a matcher could find.
Even the most careful matcher misses 40{,}949 of the 74{,}660 cross-file pairs (recall 0.452).
This number is an upper bound on the uses that are invisible to text; how much of it is due to elaboration and how much to the approximations of the matcher is not determined in this paper: the breakdown of the 40{,}949 missed cross-file pairs into instances, notation, generated declarations and other uses is left for future work.
Such uses can be measured only as the gap to the compiled index, and an estimate built only from text cannot recover them, since they are outside the scope of any capture built from text.

\paragraph{Toward an estimate that is truly compile-free}
In a setting without a compiled index, the two captures would both have to come from text, for example matcher~A and matcher~B, with the compiled index used only afterwards to judge the result.
Then the ratio of the estimate to the truth no longer reduces to \eqref{eq:lp-identity}, and false matches in either capture, positive dependence between captures that read the same text, and uses invisible to both would all enter.
We do not compute this two-text-capture estimate (matcher~A against matcher~B, per type of use, judged against the compiled index on the same 1{,}000 modules) in this paper; it is left for future work.
Our earlier study on citation graphs of textbooks found that two mechanical extractions captured almost disjoint sets of edges and gave only a conditional guideline for coverage \cite{takamatsu2026coverage}; capture--recapture has likewise been used for software inspections since the early 1990s \cite{eick1992}, at first without a known true count, and later evaluated against defects seeded into the inspected documents \cite{briand2000}, rather than against a complete record of the kind a compiler provides.

\section{Discussion}\label{sec:discussion}

This section reads the results of Sections~\ref{sec:index} and~\ref{sec:cr} together.
Each subsection states one lesson, the evidence for it in this snapshot (commit \texttt{e497a66}), and how far the lesson is expected to carry beyond it.

\subsection{False matches inflate the estimate; recall does not enter the ratio}\label{sec:disc-precision}

The Lincoln--Petersen estimator and its Chapman form assume that every item recorded by a capture is a member of the population and is identified correctly~\cite{chapman1951,chao2001}.
A text matcher violates the first assumption whenever a name in the source text is attributed to the wrong declaration, for example to a namesake in another namespace.
Such a false match increases the size $T$ of the second capture but cannot increase the overlap $M$ with the compiled index, so it moves the estimate $CT/M$ upward.
A missed match lowers $T$ and $M$ together; with the number of false matches $F=T-M>0$ held fixed, missing one correct match raises the ratio $T/M$ from $1+F/M$ to $1+F/(M-1)$ (for $M>1$). What matters is precision: at the same precision, the Lincoln--Petersen ratio is the same whatever the recall.

When the compiled index serves both as the first capture and as the ground truth, as in Section~\ref{sec:cr}, this can be stated exactly: the estimate divided by the truth $C$ is $T/M$, the reciprocal of the matcher's precision, and the matcher's recall does not enter at all.
The Chapman values follow this line closely (Fig.~\ref{fig:cr}).
From the numbers in Section~\ref{sec:cr}, $T/M$ is 1.223 for cross-file uses with the simple matcher and 1.144 with the matcher that resolves \texttt{open} and \texttt{namespace}; the reported ratios of the Chapman estimate to the truth are 1.22 and 1.14.
The improvement from the first matcher to the second is therefore an improvement in precision (81.8\% to 87.4\%), even though recall also rose (16.1\% to 45.2\%).
For same-file uses both matchers have a precision of about 97\% (96.9\% and 97.4\%), and both ratios are 1.03 although recall differs (97.3\% and 69.6\%).

The practical reading is that a text-based capture--recapture estimate should be reported together with an estimate of the precision of the matcher, obtained for instance by checking a random sample of matches, as we did for name citations in our earlier study (there with a single AI-based judge)~\cite{takamatsu2026coverage}.
Without that, the size of the estimate says more about the number of namesakes in the code than about the number of uses.
That misidentification inflates capture--recapture estimates is well known in ecology~\cite{creel2003,lukacs2005}; what our data add is a measurement of the size of the effect (which, in this design, equals $1/\text{precision}$ by construction) against a complete ground truth of 88,003 uses, rather than against seeded defects or simulated tags.

The exact identity holds only because $C$ is complete.
When neither capture is complete, positive dependence between captures (uses that are easy to see in the text are also the ones a compiler records near the use site) biases the estimate downward, and negative dependence biases it upward~\cite{chao2001}.
We observed the first kind of heterogeneity directly: with the simple matcher, recall was 0.973 for same-file uses and 0.161 for cross-file uses.
This is why the estimate was made separately for each type of use, in the same way that Corollary~1 of~\cite{takamatsu2026coverage} stratifies omission by type.

\subsection{What lies outside a capture's scope is never counted}\label{sec:disc-scope}

Capture--recapture estimates the members that at least one capture could in principle record.
A member that no capture can record has catchability zero, and no estimator built from the captures will count it.
Every capture built from text reads the same source text, so a use that exists only after elaboration (an instance found by type-class resolution, a notation, a declaration generated by a command or attribute) is outside the scope of every text capture at once.
It is not merely missed with some probability, so the overlap carries no information about it.
Section~\ref{sec:cr} measures this part as the gap between the compiled index and what the text can see.
The share of compiled uses that have no textual occurrence at all is not measured here; for cross-file uses, the 40,949 pairs missed by the matcher that resolves \texttt{open} and \texttt{namespace} give only an upper bound on it.
How these 40,949 pairs divide into instances, notations, generated declarations and other uses is not yet known and is left for future work.
The same structure appeared in our earlier study, where cross-book citations could be captured only by name and lay outside the two-list estimate~\cite{takamatsu2026coverage}.

For deletion this is the decisive point.
A declaration that is used only through an instance or a notation has no textual use anywhere and looks, to any search of the source, like a true zero.
The compiled index is the only one of our sources that records such uses, which is why we treat it as ground truth, even though it also serves as one capture.

\subsection{With small overlap the Chapman estimate is unstable}\label{sec:disc-chapman}

The Chapman estimator $\hat N=(n_1+1)(n_2+1)/(m+1)-1$ removes most of the bias of the Lincoln--Petersen estimator when the overlap $m$ is small, but its variance still grows roughly as $m^{-3}$ for fixed capture sizes~\cite{chao2001}.
In the samples of Section~\ref{sec:cr} the overlaps are large (12,025 cross-file uses with the simple matcher and 33,711 with the resolving matcher), and the sampling error is small.
That is precisely the problem: with the usual variance formula~\cite{chao2001}, the normal-approximation 95\% interval for cross-file uses with the simple matcher is 90,654 to 91,930, which is narrow and does not contain the true value 74,660.
The interval describes the sampling variability of $\hat N$ under the model's assumptions, not the bias caused by false matches, and a narrow interval should not be read as evidence that the model holds.

The opposite case arises when the estimate is broken into small units, such as single modules, namespaces or the declarations of one author.
Then $m$ may be a handful or zero, the estimate swings with a single match, and the normal approximation is poor; in our earlier study an overlap of $m=409$ already gave a standard error of 1,132 on an estimate of 26,865~\cite{takamatsu2026coverage}.
For the deletion problem, which is decided per declaration, capture--recapture is therefore not a substitute for an index: it can describe the total amount of use within a scope, but it cannot say whether a particular declaration is used.

\subsection{A rule for safe deletion}\label{sec:disc-rule}

The results of Section~\ref{sec:index} give a direct rule.
Of the \NZeroCross{} declarations with no use from other files, \NSameFileOnly{} (\PFalseZero) are used in their own file, and only \NTrueZero{} (\PTrueZero{} of \NDeclReal{} declarations) have no recorded use (Fig.~\ref{fig:falsezero}).
``No use from other files'' is thus a poor proxy for ``unused'', and the deletion in pull request \#1416, partly reverted in \#1417, is an instance of the gap: six of the eight declarations it removed were still used in their own files (Table~\ref{tab:prs}).
We therefore propose the following procedure.
\begin{enumerate}
  \item Delete a declaration only if it has no use in the compiled index, counting same-file uses and uses that arise only after elaboration.
  \item Take the index from the commit that the change will be merged into, or check that no commit since the indexed one adds a use of the declaration; in a repository where many people merge every few minutes, an index ages quickly.
  \item Build the affected modules locally before the pull request is opened, and check that the diff contains only the intended deletions.
\end{enumerate}
Step~1 addresses the failure of \#1416.
The local build of step~3 also catches uses that the compiled index cannot record, such as a definitional lemma that \texttt{simp} or \texttt{dsimp} uses without leaving a constant in the proof term (Section~\ref{sec:limitations}).
Step~3 addresses a different failure: \#1425 restored five declarations that \#1420 had removed outside its intended scope, through a change that should not have been part of that pull request.
The removal did not break the build of that file (Section~\ref{sec:index:case}), and the compiled index records no use of the five declarations from any other module, so neither a build nor an index would have caught it; only the check of the diff does.
The rule is conservative in one direction only.
It never deletes a declaration that the compiled snapshot uses, but it may keep a declaration that is used only by other declarations that are themselves unused; removing such chains requires iterating the rule or computing reachability from the main theorems, as the dependency studies of the release did~\cite{zhuk9581,yth4377,schiller3750}.

\subsection{File grouping does not explain the false zeros}\label{sec:disc-partition}

Whether a use counts as cross-file or same-file depends not only on the use relation but also on how declarations are grouped into files.
An analogy suggested by T.~Akiyama makes this concrete: when two departments merge, the same two people making the same call are recorded, overnight, as making an internal call instead of an inter-departmental one.
Pull request \#120 regrouped the declarations of the release into 47,885 modules, so it was natural to ask whether the false zeros of Fig.~\ref{fig:falsezero} are such relabelled calls.
Section~\ref{sec:3-filegrouping} answers this question with the test T.~Akiyama suggested: with the use relation of the compiled index held fixed and only the file partition set back to that before the regrouping, the share of false zeros on the 99.47\% of declarations that can be mapped is 96.73\% under the current partition and 96.72\% under the earlier one, and only 211 of the 437,349 false zeros (0.048\%) were used from another file under the earlier partition.
In the terms of the analogy, the calls were already internal before the departments merged.
The false zeros come instead from auxiliary lemmas and definitions that already sat in the same file as the declarations that use them before the regrouping.
The rule of Section~\ref{sec:disc-rule} does not depend on this result, because it counts uses regardless of which file they come from.

\subsection{Relation to repeated fixes}\label{sec:disc-repeat}

Ambiguous names connect the deletion problem to a second pattern in the history of the same branch.
In the commit log of \texttt{opt/collapse-and-consolidate-4.36} up to 25 November 2026, 06:13 UTC (1,404 commits, about twelve hours before the snapshot), 141 files were fixed two or more times by commits touching at most five files.
In 81 of the 204 consecutive fix pairs on those files (39.7\%), the later fix deleted a line the earlier fix had added or restored a line it had removed.
This is an upper bound on actual undoing, since a line that was merely improved also counts; 12 of the 81 later commits contain ``restore'', ``revert'' or ``original'' in their messages.
Most of the back-and-forth we inspected was a choice between two declarations with the same name, where one fixer qualified a reference to one namesake and the next switched it to the other.

The compiled index bears on this pattern in two ways.
First, among 679 name-qualification fixes, in 17 (2.5\%) to 90 (13.3\%) of them every other namesake is unused in the current index; deleting the unused copy would have removed the ambiguity at its source instead of choosing between the copies again.
These counts use the index of a later commit than the fixes and are approximate.
Second, several of the first author's pull requests (Table~\ref{tab:prs}) remove unused declarations that collide with used namesakes, which is the same remedy applied before rather than after a ping-pong.
Both observations point in the same direction as the main result: decisions about use, whether to delete a declaration or which namesake to refer to, are more stable when they are taken from the compiled index than when they are taken from a search of the text.

\section{Limitations, availability, and use of generative AI}\label{sec:limitations}

\subsection{Limitations}

\paragraph{A single snapshot}
All results refer to one commit, e497a66 (25 November 2026, 17:55 UTC), of branch \texttt{opt/collapse-and-consolidate-4.36} of \texttt{rohanarun/riemann-hypothesis}.
This is the last commit on that branch: merging stopped there, and community work moved to \texttt{leanprover-community/riemann-hypothesis}.
The snapshot makes the counts reproducible, but it also fixes them in time.
The shares in Fig.~\ref{fig:falsezero} describe one stage of an ongoing consolidation, and we have not measured how they change as files are merged, split or pruned further in the successor repository.
In particular, whether a use counts as ``same-file'' or ``cross-file'' depends on how declarations are grouped into files, not only on the use relation itself.
As suggested by T.~Akiyama, we separated this effect for one earlier grouping, the file partition before consolidation (commit 40e56bab02, 174,793 \texttt{.lean} files): with the use relation held fixed, the share of false zeros was 96.73\% under the current partition and 96.72\% under the pre-consolidation partition, on the same 452,153 declarations with no cross-file use whose earlier file could be traced (99.47\% of the \NZeroCross{}; Section~\ref{sec:3-filegrouping}).
Only 211 of the 437,349 false zeros of this subset (0.048\%) had a cross-file use under the earlier partition; counting every undecided case as cross-file gives at most 931 (0.21\%), and the largest upper bound over all mappings is 0.51\%. Consolidation therefore does not explain the share.

\paragraph{Module coverage of the compiled index}
The compiled index covers \NModCovered{} of the \NModTotal{} modules of the snapshot; the other 105 modules declare nothing of their own (104 import-only modules and the statement file \texttt{Challenge} with three definitions), so no declaration of the proof is missing from the counts. The \texttt{Challenge} definitions themselves are not in the index.
Uses that originate in modules outside the index are not recorded.
Missing uses can only remove links, never add them, so the counts of declarations with no cross-file use (\NZeroCross{}) and with no recorded use at all (\NTrueZero{}) are upper bounds for the declarations of the covered modules.
This is the situation analysed, under independent omission of links, in Proposition~1 of our method paper~\cite{takamatsu2026coverage}; here omission is by module rather than independent, but the direction is the same: incomplete forward links produce false ``never cited'' counts, and items with a single true citation are the most exposed.
A declaration counted in \NTrueZero{} should therefore be read as a candidate for deletion, not as a verified unused declaration, until the build has been checked.

\paragraph{Mapping declarations to pre-consolidation files}
The earlier file of each declaration was found by parsing the source of commit 40e56bab02 without compiling it, and was checked against the \texttt{-- bundled from} markers in the source of the snapshot; of the declarations mapped by both methods, 99.985\% were mapped to the same earlier file, and only 60 of the 120,637 earlier files with declarations (0.05\%) were split across current modules.
The parser only approximates Lean's naming, for example for unnamed instances, and the mapped subset leaves out 0.53\% of the declarations with no cross-file use, the largest group being auxiliary names of macros and notation.
The use relation was taken from the snapshot, so declarations removed or renamed between 40e56bab02 and e497a66 are not represented, among them the thin wrappers in the 53,966 files deleted by consolidation.
Counting the names that occur in those deleted files as cross-file uses (matching the last two components of a name), 2,843 false zeros (0.65\%) would be attributed to consolidation and the deleted wrappers, and the share would be 96.83\%; this sensitivity check is based on names only, not on compiled uses.

\paragraph{What the compiled index can and cannot see}
The index records a use when a declaration appears as a constant in the compiled form of another declaration.
Auxiliary constants generated by the elaborator (for example equation lemmas, match and recursor helpers, constructor helpers, and the auxiliary constants under private declarations) were folded into the user-written declaration they belong to by a fixed list of name patterns; this mapping is a heuristic on names, and \NDeclReal{} depends on it.
With a narrower rule the share of false zeros is 95.8\% and with a broader rule 96.7\%, so the main claim does not depend on the rule.
Private declarations written by users remain real declarations; they account for 60,349 of the \NSameFileOnly{} false zeros (13.8\%).
Uses that leave no constant in a compiled term, such as references made only from metaprograms, commands that are not part of the proof, or code in repositories that build on the snapshot, are outside the index.
The index can also miss uses made at the source level: a definitional lemma (proved by \texttt{rfl}) that is used by \texttt{simp} or \texttt{dsimp} leaves no constant in the elaborated proof term, so a declaration with no recorded use can still be needed.
In trial deletions on a later version of the proof (\texttt{leanprover-community/riemann-hypothesis}), such deletions broke \texttt{simp} calls in the same or in other files.
This is why the deletion procedure includes a local build (step~3 in Section~\ref{sec:disc-rule}).
For this reason ``no use in the compiled index'' is a necessary condition for safe deletion, and a successful build after deletion remains the final check.

\paragraph{Choice of text matchers}
The capture-recapture results in Section~\ref{sec:cr} describe three specific text matchers built from regular expressions: (A) a match on the last two components of a qualified name, together with short names that occur at least twice in their own file; (B) a match after resolving \texttt{open} and \texttt{namespace} prefixes; and (C) matcher B together with short names in the same file.
None of them is a Lean parser.
In particular, matcher B collects the \texttt{open} and \texttt{namespace} prefixes over the whole file rather than over their exact scope, which can add candidates that the elaborator would not accept.
Other designs, such as a matcher based on the Lean syntax tree, would give other values of precision and recall, and hence other ratios of estimate to truth.
What we expect to carry over is the structural relation, not the particular values: when the compiled index serves both as one capture and as ground truth, the Lincoln--Petersen estimate divided by the truth equals $T/M$, the reciprocal of the matcher's precision (Section~\ref{sec:cr} and Fig.~\ref{fig:cr}).
The same identity means that this part of the study is not an independent test of a two-list design in which neither list is complete; it isolates the effect of the text matcher.

\paragraph{A sample of 1,000 modules}
The capture-recapture comparison uses 1,000 modules drawn at random (seed 20261126) from the \NModCovered{} modules of the compiled index, with 88,003 compiled uses as ground truth.
We drew one sample only, so the sampling variability of the reported ratios is not quantified, and we report no interval for them.
As in Section~\ref{sec:cr}, the unit of counting is a distinct pair of a using module and a used declaration.
Modules differ greatly in size and in the number of \texttt{open} commands, and a stratified sample could give different precision and recall for the matchers; we have not tested this.

\paragraph{Pull requests as a case study}
The pull requests in Table~\ref{tab:prs} were opened by one of the authors on the snapshot branch.
They illustrate how a deletion broke the build and had to be undone in part (\#1416 $\rightarrow$ \#1417, which restored six of the eight removed declarations that were still used in their own files; the same-file check of \#1416 was a text search), how declarations removed outside the intended scope of a pull request were restored (\#1420 $\rightarrow$ \#1425, which restored five declarations removed by \#1420), and how deletions based on the compiled index were accepted.
On the snapshot branch, 18 pull requests of the first author were merged (\#1415--\#1433; \#1423 is by another contributor): 16 deletions, including \#1416, and the 2 restorations \#1417 and \#1425. In addition, 14 pull requests (\#1434--\#1449, excluding \#1440 and \#1441, which are not the first author's) remained open when merging on the branch stopped.
The second case was a failure of the tooling, not of the deletion rule: an edit to another file, for which no pull request was opened because its candidates failed a re-check, was left in the working tree and carried into \#1420 (Section~\ref{sec:index:case}).
As far as we have checked, \#1420 did not break the build. On 27 November 2026 we compiled the file \texttt{IsUnitaryOfOrder.lean} as of the merge commit 94167b2 (after \#1420), and the version before \#1420, with Lean; both compiled without errors. Both were compiled against the dependencies of the snapshot e497a66, not those of 94167b2.
The complete compiled index records no use of the five removed declarations (\texttt{GlobalTheory.Mem.Chr.etaAlg}, \texttt{Mu0Exists}, \texttt{HChar.resFin\_apply}, \texttt{HChar.resInf\_apply}, \texttt{isUnitAt\_nat}) from any other module.
The only same-file uses are of the auxiliary proof constant \texttt{etaAlg.\_proof\_1} by \texttt{DeltaExists}, \texttt{EtaTExists} and \texttt{Mu0Exists}. Lean shares an identical proof term among declarations by abstracting it as an auxiliary constant under the first declaration that needs it; when \texttt{etaAlg} is removed, the term is abstracted under another declaration, so these are not genuine dependencies on \texttt{etaAlg}.
\#1425 was therefore a restoration of declarations removed outside the intended scope of \#1420, not a repair of a broken build.
These pull requests are an observational record of a small number of changes, not a controlled comparison of deletion rules, and acceptance by the maintainers depended also on review practice and timing.
One of the authors is himself a contributor to the repository studied, which should be kept in mind when reading this part.

\subsection{Data availability}
The input is public: the repository and commit named above identify it completely.

\subsection{Use of generative AI}
The pull requests discussed in this paper, the scripts that build the index, run the capture-recapture comparison and recompute the shares under the pre-consolidation partition, and the analyses were made with Claude Code using the model Claude Opus 5.5 (Anthropic; November 2026), working under the direction of the authors.
All 32 commits of the first author's pull requests carry the trailer ``Co-Authored-By: Claude Opus 5.5''.
The manuscript was also drafted with Claude Opus 5.5.
T.~Akiyama used OpenAI Codex to recheck equation~\eqref{eq:lp-identity} and Table~\ref{tab:cr}.
The numbers reported here were taken from the outputs of the scripts, not from text generated by the AI system.
No generative AI system is an author, and the authors take full responsibility for the content of this paper, including the pull requests submitted to the community repository.

\section*{Acknowledgment}
We thank Fumihiko Nakamura (Kitami Institute of Technology), Yuri Yatagawa (Institute of Science Tokyo), and Shun'ichi Honda (Chitose Institute of Science and Technology) for discussions and comments, and Hibiki Ito (University of Helsinki and Kyoto University) for support.

\bibliographystyle{IEEEtran}
\bibliography{refs_v16}
\end{document}